# Automatic Spatially-aware Fashion Concept Discovery

Xintong Han[1,3]  Zuxuan Wu[1]  Phoenix X. Huang[2]  Xiao Zhang[3]
Menglong Zhu[3]  Yuan Li[3]  Yang Zhao[3]  Larry S. Davis[1]
[1]University of Maryland  [2]Snap Inc.  [3]Google Inc.
{xintong,zxwu,lsd}@umiacs.umd.edu  phoenix@snap.com
{andypassion,menglong,liyu,yzhao}@google.com

## Abstract

*This paper proposes an automatic spatially-aware concept discovery approach using weakly labeled image-text data from shopping websites. We first fine-tune GoogleNet by jointly modeling clothing images and their corresponding descriptions in a visual-semantic embedding space. Then, for each attribute (word), we generate its spatially-aware representation by combining its semantic word vector representation with its spatial representation derived from the convolutional maps of the fine-tuned network. The resulting spatially-aware representations are further used to cluster attributes into multiple groups to form spatially-aware concepts (e.g., the neckline concept might consist of attributes like v-neck, round-neck, etc). Finally, we decompose the visual-semantic embedding space into multiple concept-specific subspaces, which facilitates structured browsing and attribute-feedback product retrieval by exploiting multimodal linguistic regularities. We conducted extensive experiments on our newly collected Fashion200K dataset, and results on clustering quality evaluation and attribute-feedback product retrieval task demonstrate the effectiveness of our automatically discovered spatially-aware concepts.*

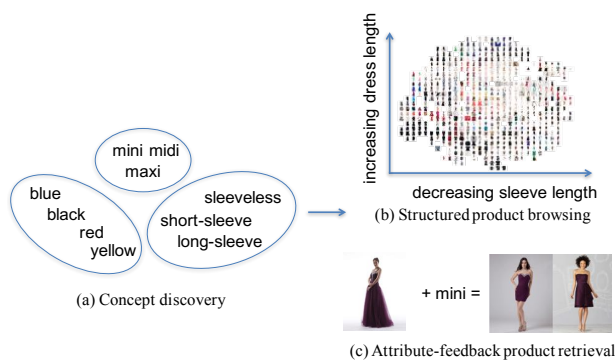

Figure 1. (a) We propose a concept discovery approach to automatically cluster spatially-aware attributes into meaningful concepts. The discovered spatially-aware concepts are further utilized for (b) structured product browsing (visualizing images according to selected concepts) and (c) attribute-feedback product retrieval (refining search results by providing a desired attribute).

## 1. Introduction

The exponential growth of online fashion shopping websites has encouraged techniques that can effectively search for a desired product from a massive collection of clothing items. However, this remains a particularly challenging problem since, unlike generic objects, clothes are usually subject to severe deformations and demonstrate significant variations in style and texture, and, most importantly, the long-standing semantic gap between low-level visual features and high-level intents of customers is very large. To overcome the difficulty, researchers have proposed interactive search to refine retrieved results with humans in the loop. Given candidate results, customers can provide various feedback, including the relevance of displayed images [20, 4], or tuning parameters like color and texture, and then results are updated correspondingly. However, relevance feedback is limited due to its slow convergence to meet the customer requirements. In addition to color and texture, customers often wish to exploit higher-level features, such as *neckline*, *sleeve length*, *dress length*, *etc*.

Semantic attributes [13], which have been applied effectively to object categorization [15, 27] and fine-grained recognition [12] could potentially address such challenges. They are mid-level representations that describe semantic properties. Recently, researchers have annotated clothes with semantic attributes [9, 2, 8, 16, 11] (*e.g.*, material, pattern) as intermediate representations or supervisory signals to bridge the semantic gap. However, annotating semantic attributes is costly. Further, attributes conditioned on object parts have achieved good performance in fine-grained recognition [3, 33], confirming that spatial information is critical for attributes. This also holds for clothing images. For example, the *neckline* attribute usually corresponds to the top part in images while the *sleeve* attribute ordinarily

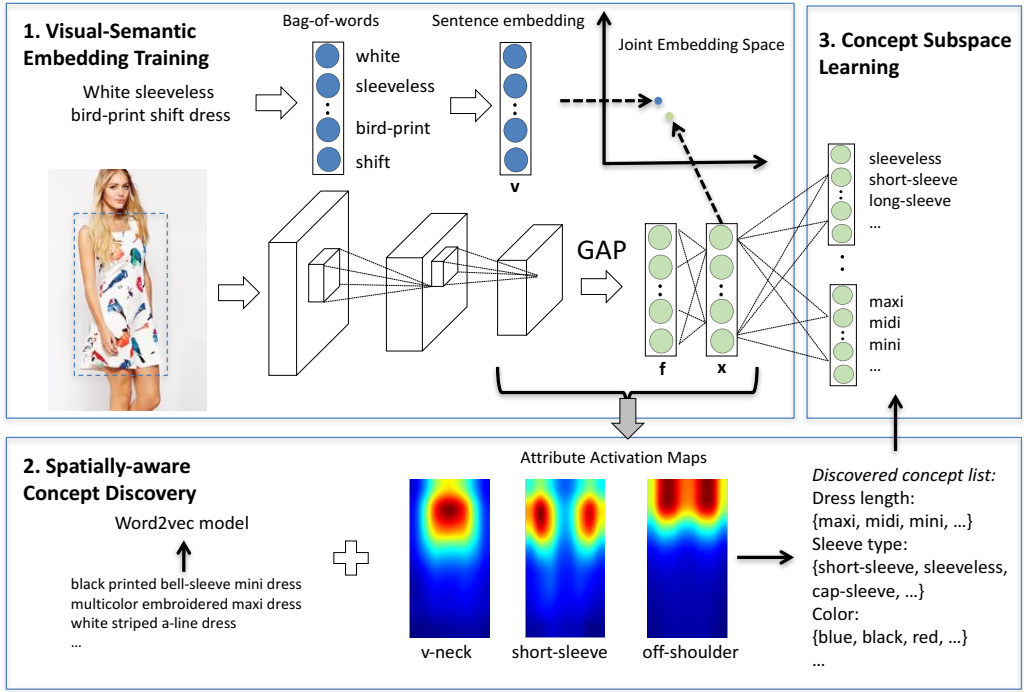

Figure 2. Overview of our approach. Our approach mainly contains three parts: 1. Joint embedding space training. A joint visual-semantic embedding space is trained using clothing images and their product descriptions. 2. Spatially-aware concept discovery. We use neural activations provided by global pooling (GAP) layer to generate attribute activation maps (AAMs) of attributes. The AAM captures the spatial information of attributes (*i.e.*, what is the spatial location an attribute usually refers to). By combining attributes' spatial information and their semantic representations obtained from a word2vec model, we cluster attributes into concepts. 3. Concept subspace learning. For each discovered concept, we further train a sub-network to effectively measure the similarity of images according to this concept only.

relates to the left and right side of images.

To address the above limitations, we jointly model clothing images and their product descriptions with a visual-semantic embedding, and propose a novel approach that automatically discovers spatially-aware concepts, each of which is a collection of attributes describing the same characteristic (*e.g.*, if the concept is *color* then the attributes could contain *yellow* and *blue*, as shown in Figure 1(a)). In addition, we learn a subspace embedding for each discovered concept to facilitate a structured exploration of the dataset based on the concept of interest (Figure 1(b)). More importantly, inspired by [10], we leverage the learned visual-semantic space to exploit multimodal linguistic regularities for attribute-feedback product retrieval. For example, an image of a "white sleeveless dress" − "sleeveless" + "long-sleeve" would be near images of "white long-sleeve dress". In contrast to [10] which requires explicitly specifying the attribute to remove, we implicitly remove corresponding attributes based on the discovered concepts (Figure 1(c)).

Figure 2 provides an overview of the framework. Specifically, our framework contains the following three steps (1) we first train a joint visual-semantic embedding space using clothing images and their product descriptions. Given an image, we compute its features with GoogleNet, which are further projected into the embedding space to minimize the distance to its product description encoded by bag-of-words of attributes. By fine-tuning GoogleNet in an end-to-end fashion, we train a discriminative model that contains localization information of attributes; (2) we then obtain the spatial representation for each attribute, indicating where in images the attribute mostly corresponds to, from the attribute activation maps. These spatial representations are further utilized to augment their corresponding semantic word representations (word vectors) produced from a skip-gram model. Further, clustering is performed to discover concepts, each of which contains semantically related attributes (*e.g.*, *maxi*, *midi*, *mini* are all different *dress length*); (3) we further disentangle the trained visual-semantic embedding by training a subspace embedding for each discovered concept, in which the similarities among items can be measured based on the corresponding concept only. The transformation of images into a subspace embedding facilitates attribute-feedback clothing search and structured browsing of fashion images.

Given the fact that existing datasets only contain images and annotated attributes (which are often very sparse) rather than image and product description pairs, we con-structed the Fashion200K dataset, which contains more than 200,000 clothing images of five categories (dress, top, pants, skirt and jacket) and their associated product descriptions from online shopping websites. These five classes are the most important verticals in fashion due to their various styles and occasions. Thus, we focus on these categories in our dataset, but our method is applicable to any fashion categories. We conduct extensive experiments on this dataset to validate the efficacy of the automatically discovered concepts in attribute-feedback product retrieval as well as structured fashion image browsing.

Our main contributions are two-fold. First, we demonstrate that the augmentation of semantic word vectors for attributes with their spatial representations can be used to effectively cluster attributes into semantically meaningful and spatially-aware concepts. Second, we leverage semantic regularities in the visual-semantic space for attribute-feedback clothing retrieval.

## 2. Related Work

**Interactive image search**. Extensive studies have been conducted on interactive image search, aiming to improve retrieved results from search engines with user feedback [20, 11, 4] (See [35] for a comprehensive review). The basic idea is to refine the results by incorporating feedback from users, including the relevance of the candidates, and tuning low-level parameters like color and texture. In practice, relevance feedback requires a large number of iterations to converge to user intent. Also, it requires manual annotations to define the relative attributes, which limits its scalability. In addition, when searching clothing images, customers generally focus on certain higher-level characteristics, such as *neckline*, *sleeve length*, *etc*., thus rendering relevance feedback less useful.

**Attributes for clothing modeling**. There have been numerous works focusing on utilizing semantic attributes as mid-level representations for clothing modeling. For instance, Chen *et al*. [2] learned semantic attributes for clothing on the human upper body. Huang *et al*. [8] built tree-structured layers for all attribute categories to form a semantic representation for clothing images. Veit *et al*. [29] learned visually relevant semantic subspaces using a multi-query triplet network. Kovashka *et al*. [11] utilized relative attributes with ranking functions instead of using binary feedback for retrieval tasks. In contrast, we propose a novel concept discovery framework, in which a concept is a collection of automatically identified attributes derived by jointly modeling image and text.

**Visual concept discovery**. To exploit the substantial amounts of weakly labeled data, researchers have proposed various approaches to discover concepts. Sun *et al*. [24] combined visual and semantic similarities of concepts to cluster concepts while ensuring their discrimination and compactness. Vittayakorn *et al*. [30] and Berg *et al*. [1] verified the visualness of attributes, and [30] also uses neural activations to learn the characteristics of each attribute. Vaccaro *et al*. [28] utilized a topic model to learn latent concepts and retrieve fashion items based on textual specifications. Singh *et al*. [22] discovered pair-concepts for event detection and discard irrelevant concepts by the co-occurrences of concepts. Recently, some works discovered the spatial extents of concepts. Xiao and Lee [32] discovered visual chains for locating the image regions that are relevant to one attribute. Singh and Lee [23] introduced a deep network to jointly localize and rank relative attributes. However, these approaches involve training a single model for each individual attribute, which is not scalable.

**Visual-semantic joint embedding**. Our work is also related to visual-semantic embedding models [5, 10, 31, 14, 7]. Frome *et al*. [5] recognize objects with a deep visual-semantic embedding model. Kiros *et al*. [10] adopted an encoder-decoder framework coupled with a contrastive loss to train a joint visual-semantic embedding. Wang *et al*. [31] combined cross-view ranking loss and within-view structure preservation loss to map images and their descriptions. Beyond training a joint visual-semantic embedding with image and text pairs as in these works, we further decompose the trained embedding space into multiple concept-specific subspaces, which facilitates structured browsing and attribute-feedback product retrieval by exploiting multimodal linguistic regularities.

## 3. Fashion200K Dataset

There have been several clothing datasets collected recently [16, 8, 21, 6, 7]. However, none of these datasets are suitable for our task because they do not contain descriptions of images. This prevents us from learning semantic representations for attributes using word2vec [18]. Thus, we collected the Fashion200K dataset and automatically discover concepts from it.

We first crawled more than 300,000 product images and their product descriptions from online shopping websites and removed the ones whose product descriptions contain fewer than four words, resulting in over 200,000 images. We then split them into 172,049 images for training, 12,164 for validation, and 25,331 for testing. For cleaning product descriptions, we deleted stop words, symbols, as well as words that occur fewer than 5 times. Each remaining word is regarded as an attribute. Finally, there are 4,404 attributes for training the joint embedding.

Example clothing image and description pairs are shown in Figure 3. Since we wish to automatically discover concepts from this noisy dataset and learn concept-level subspace features, we do not conduct any manual annotations for this dataset. Note that as a preprocessing step, we trained a detector using the MultiBox model [25] for all five

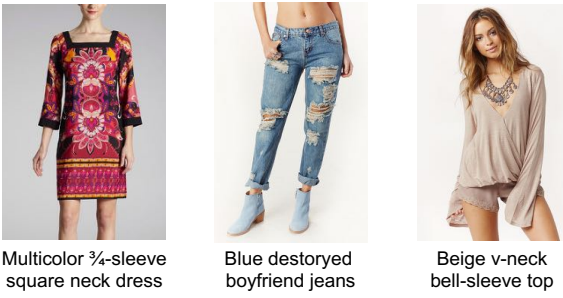

Multicolor ¾-sleeve square neck dress | Blue destoryed boyfriend jeans | Beige v-neck bell-sleeve top

Figure 3. Examples of the image-text pairs in Fashion200K.

categories and run them on all images. Only the detected foregrounds are cropped and used as input to our model.

## 4. Our Approach

In this section, we present the key components of the proposed concept discovery approach shown in Fig. 2, including visual-semantic embedding learning, spatially-aware concept discovery and concept subspace learning. Since our method leverages spatial information of an attribute, and the same attribute in different types of clothing (*e.g.*, "short" in "short dress" and "short pants") will have different spatial characteristics, we train an individual model for each category in our dataset. For simplicity in notation and illustration, we only show the concept discovery approach for dresses, while the same pipeline is applied to other categories in the same fashion. Results of all categories are shown and evaluated in our experiments.

### 4.1. Visual-semantic Embedding

To fully explore the substantial weakly labeled web data for mining concepts, we first train a joint visual-semantic embedding model with image-text pairs by projecting a product image and its associated text into a joint embedding space. Following [10], we also utilize a stochastic bidirectional contrastive loss to achieve good convergence.

More formally, let $\mathbf{I}$ denote an image and $S = \{w_1, w_2, ..., w_N\}$ its corresponding text, where $w_i$ is the $i$-th attribute (word) in the product description. Let $\mathbf{W_I}$ denote the image embedding matrix, and $\mathbf{W_T}$ denote the attribute embedding matrix. We first represent the $i$-th word $w_i$ with one-hot vector $\mathbf{e}_i$, which is further encoded into the embedding space by $\mathbf{v}_i = \mathbf{W_T} \cdot \mathbf{e}_i$. We then represent the product description with bag-of-words $\mathbf{v} = \frac{1}{N}\sum_i \mathbf{v_i}$. Similarly, for the image $\mathbf{I}$, we first compute its feature vector $\mathbf{f} \in \mathbb{R}^{2048}$ with a GoogleNet model [26] parameterized by weights $\mathbf{V}$ after the global average pooling (GAP) layer as shown in Figure 2. Then we project the feature vector into the embedding space, in which the original image is represented as $\mathbf{x} = \mathbf{W_I} \cdot \mathbf{f}$.

The similarity between an image and its description is computed with *cosine* similarity, *i.e.*, $d(\mathbf{x}, \mathbf{v}) = \mathbf{x} \cdot \mathbf{v}$, where $\mathbf{x}$ and $\mathbf{v}$ are normalized to unit norm. Finally, the joint embedding space is trained by minimizing the following contrastive loss:

$$\min_{\Theta} \sum_{\mathbf{x},k} \max(0, m - d(\mathbf{x},\mathbf{v}) + d(\mathbf{x},\mathbf{v}_k)) + \sum_{\mathbf{v},k} \max(0, m - d(\mathbf{v},\mathbf{x}) + d(\mathbf{v},\mathbf{x}_k)), \quad (1)$$

where $\Theta = \{\mathbf{W_I}, \mathbf{W_T}, \mathbf{V}\}$ contains the parameters to be optimized, and $\mathbf{v}_k$ denotes non-matching descriptions for image $\mathbf{x}$ while $\mathbf{x_k}$ are non-matching images for description $\mathbf{v}$. By minimizing this loss function, the distance between $\mathbf{x}$ and its corresponding text $\mathbf{v}$ is forced to be smaller than the distance from unmatched descriptions $\mathbf{v_k}$ by some margin $m$. Vice versa for description $\mathbf{v}$.

### 4.2. Spatially-aware Concept Discovery

The training process of a joint visual-semantic embedding will lead to a discriminative CNN model, which contains not only the semantic information (*i.e.*, the last embedding layer) but also important spatial information that is hidden in the network. We now discuss how to obtain spatially-aware concepts from the network.

**Attribute spatial representation**. Spatial information of an attribute is crucial for understanding what part of a clothing item the attribute refers to. Motivated by [34], we generate embedded attribute activation maps (EAAM), which can localize the salient regions of attributes for an image by a single forward pass with the trained network.

Given an image $\mathbf{I}$, let $q_k(i,j)$ be the activation of unit $k$ in the last convolutional layer at location $(i,j)$. After the global average pooling (GAP) operation, $f_k = \sum_{i,j} q_k(i,j)$ is the $k$-th dimension feature of the image representation $\mathbf{f}$. For a given attribute $a$, the cosine distance $d(\mathbf{x}, \mathbf{W}^a)$ between image embedding $\mathbf{x}$ and attribute embedding $\mathbf{W}^a$ indicates the probability that attribute $a$ is present in this image. If we plug $f_k$ into the cosine distance we obtain:

$$\begin{aligned} d(\mathbf{x}, \mathbf{W}^a) &= \sum_m W^a_m x_m = \sum_m W^a_m \sum_k W_{I_{m,k}} f_k \\ &= \sum_m W^a_m \sum_k W_{I_{m,k}} \sum_{i,j} q_k(i,j) \\ &= \sum_{i,j} \sum_m W^a_m \sum_k W_{I_{m,k}} q_k(i,j) \end{aligned} \quad (2)$$

where $W^a_m$ and $W_{I_{m,k}}$ are entries of the attribute embedding $\mathbf{W}^a$ and image embedding matrix $\mathbf{W_I}$, respectively. Thus, the embedded attribute activation map (EAAM) for attribute $a$ of image $\mathbf{I}$ can be defined as:

$$\mathbf{M}^{\mathbf{I}}_a(i,j) = \sum_m W^a_m \sum_k W_{I_{m,k}} q_k(i,j) \quad (3)$$

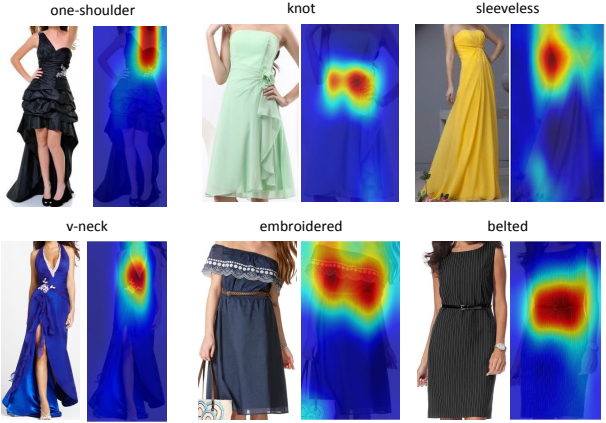

Figure 4. Embedding attribute activation map for a given attribute. The generated activation maps successfully highlight the discriminative regions for the given attribute.

Since $d(\mathbf{x}, \mathbf{W}^a) = \sum_{i,j} \mathbf{M}_a^{\mathbf{I}}(i,j)$, $\mathbf{M}_a^{\mathbf{I}}(i,j)$ indicates how likely the attribute appears at spatial location $(i,j)$.

Figure 4 shows sample EAAMs of images. We can see the activation maps indicate where the joint embedding model looks to identify an attribute. Product images on shopping websites usually have clean backgrounds and are displayed in an aligned frontal view. Thus, for a particular attribute $a$ and its positive training set (*i.e.*, images whose product descriptions contain $a$) $P_a$, we average EAAMs for all images in $P_a$ to generate an activation map $A_a$. We refer to it as the attribute activation map (AAM) of $a$:

$$\mathbf{A}_a = \frac{1}{|P_a|} \sum_{\mathbf{I} \in P_a} \mathbf{M}_a^{\mathbf{I}} \quad (4)$$

Figure 5 shows AAMs of some attributes for the dress category. From this figure, we can discover that for attributes that have clear spatial information in a dress image, their AAMs capture the spatial patterns. For example, *belt* is most likely to occur in the middle part of dress images, *long-sleeve* often occurs on two sides of dress images, and *off-shoulder* is around the shoulder region of a dress. However, for some attributes whose locations are not certain for different dress images, like *floral*, *stripe*, and colors, their AAMs span almost the entire image.

Therefore, for each attribute in a clothing category, its AAM can serve as a spatial representation. If two attributes describe the similar spatial part of a clothing category, *e.g.*, *sleeveless* and *long-sleeve*, or *v-neck* and *mockneck*, their spatial information should also be similar.

**Attribute semantic representation**. Only using spatial information is not sufficient for effective concept discovery, especially for those attributes that do not have a discriminative spatial representation. Thus, we train a skip-gram model [18] on the descriptions of clothing products to obtain the semantic representations (Word2vec vectors) for all attributes in our dataset. We denote the semantic representation of attribute $a$ as $\mathbf{E}_a$.

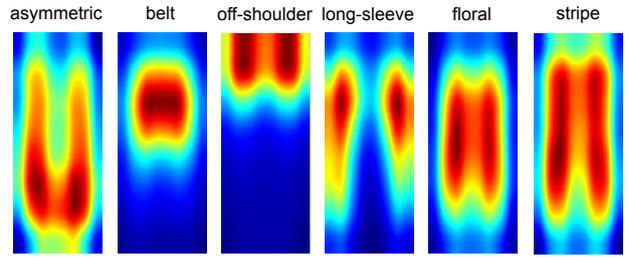

Figure 5. Attribute activation map for a given attribute of the dress category. The most frequency locations an attribute corresponds to in an image are highlighted.

| | concepts discovered by our method |
|---|---|
| dress | *dress length*: maxi, midi, mini<br>*neckline*: v, plunge, deep, high, scoop<br>*shoulder*: off-the-shoulder, one-shoulder, strapless, ... |
| top | *decoration*: lace, embellished, embroidered, beaded, ...<br>*sleeve length*: sleeveless, long-sleeve, short-sleeve, ...<br>*sleeve shape*: kimono, cap, dolman, bell, flutter, ... |
| pants | *color*: black, blue, multicolor, gray, white, green, ...<br>*pant cut*: straight-leg, slim-leg, tapered-leg, bootcut, ...<br>*pattern*: check, geometric, leopard, palm, abstract, ... |

Table 1. Concept discovered by our method. Each row contains the attributes belong to one concept. Ellipsis is used when the attribute list is too long to show.

**Attribute clustering**. Ideally, attributes belonging to the same concept describe the same characteristic of a clothing category; that means they should be both spatially consistent and semantically similar. Thus, for an attribute $a$, by simply flattening its spatial representation $\mathbf{A}_a$ and concatenating it with its semantic representation $\mathbf{E}_a$, we can generate a feature vector:

$$\mathbf{F}_a = [vec(\mathbf{A}_a), \mathbf{E}_a] \quad (5)$$

where $vec(\cdot)$ is vectorization operation and we normalize $vec(\mathbf{A}_a)$ and $\mathbf{E}_a$ to have unit norm before concatenation. As a result, this attribute feature is aware of the spatial information of the attribute and can also capture its semantic meaning. K-means clustering algorithm is then used to cluster all the attributes into attribute groups, such that the attributes within a group form a concept. Unlike [24], we do not directly use visual similarity between attributes because attributes describing the same characteristic might be visually dissimilar. For example, *blue* and *red* are both color attributes, but they are visually very different.

Table 1 presents some concepts discovered by our method for different categories. We find that the attributes describing the same characteristic are grouped into one cluster. For example, all attributes describing colors are in one concept because they are very close in the semantic embedding space (they are often the first word in product descriptions) and their AAMs do not provide much useful information (the right two AAMs in Figure 5). Thus, the semantic representations of those attributes dominate in this case and place them in the same concept. Different kinds of sleeves also form a concept, since their AAMs are very similar (along with the two sides of dresses or tops) and their word vectors are also close. We also observe that our method can successfully group noisy (not visually perceptible) attributes together, because the semantic and spatial information of these attributes is not discriminative. These noisy clusters will be discovered by our method and not affect the attribute-feedback, since customers will not provide an attribute with low visualness for retrieval.

### 4.3. Concept Subspace Learning

The discovered concepts are further utilized to refine the learned joint visual-semantic space, so that similarities between items can be measured by each individual concept (*e.g.*, *color* and *neckline* could result in different similarities). This is crucial for cases when customers want to modify attributes to refine the search results or hope to browse products based on a particular concept. Therefore, given the concepts discovered by the attribute clustering process, we further train a sub-network for each concept, constructing a concept-specific subspace.

For a concept $C = \{a_1, a_2, ..., a_n\}$ where $a_i$ is an attribute in this concept, we build a fully-connected layer and a softmax layer on top of the image embedding features to classify the $a_i$. The number of neurons in the softmax layer is $n + 1$ (each attribute corresponds to one neuron with an additional one for none-of-above). This network is trained only on images with $a_i$ in their product descriptions plus a small number of randomly sampled negative images. We denote $\mathbf{S}^C(\mathbf{x})$ to be the softmax output of the sub-network for concept $C$ given the input image $\mathbf{x}$.

After the subspace training stage, the concept subspace features (hidden layer representations) are aware of the attributes of this particular concept, and hence enable the similarity measurement among images based only on this concept. For example, a "blue maxi dress" is more similar to a "blue mini dress" than a "red maxi dress" in the *color* subspace. However, a "red maxi dress" is closer to "blue maxi dress" in *dress length* subspace. As a result, customers can choose the desired similarity measure during online shopping so they can better explore the clothing gallery.

### 4.4. Attribute-feedback Product Retrieval

Based on the discovered concepts and learned concept subspaces, we leverage multimodal linguistic regularities to help perform attribute-feedback product retrieval task. Some example results can be found in Figure 7.

Given a retrieved image ("red sleeveless mini dress", for example), customers may want to change one attribute of the image while keeping others fixed, say "I want this dress to have long-sleeves". As we already trained a visual-semantic embedding (VSE), a baseline method would be sorting database images based on their cosine distances with the query image + query attribute (*long-sleeve*). In this way, the retrieved images have a high score for the query attribute and are similar to the query image at the same time. For a query image $\mathbf{x_q}$ and a query attribute $\mathbf{w_p}$, the attribute-feedback retrieval task to find image $\mathbf{x_o}$ is defined as:

$$\mathbf{x_o} = \arg\max_{\mathbf{x}} \; (\mathbf{x_q} + \mathbf{w_p}) \cdot \mathbf{x} \tag{6}$$

However, one problem with this approach is that it retrieves images which are closest to "red sleeveless long-sleeve mini dress" instead of "red long-sleeve mini dress". To overcome this, we note that by providing a query attribute, customers implicitly intend to remove an existing attribute (*sleeveless* in this case) that describes the same characteristic of the product as the query attribute. Since the attributes within one discovered concept describe the same characteristic, we detect the implicit negative attribute $\mathbf{w_n}$ and use it to search image $\mathbf{x_o}$:

$$\begin{aligned}\mathbf{w_n} &= \arg\max_{\mathbf{w} \in C} \; \mathbf{S}^C(\mathbf{x_q}) \\ \mathbf{x_o} &= \arg\max_{\mathbf{x}} \; (\mathbf{x_q} + \mathbf{w_p} - \mathbf{w_n}) \cdot \mathbf{x}\end{aligned} \tag{7}$$

where $C$ is the concept to which $\mathbf{w_p}$ belongs and $\mathbf{S}^C(\mathbf{x_q})$ is the softmax output of the sub-network for $C$. Thus, $\mathbf{w_n}$ is the attribute in $C$ that is most likely to be present in the query image $\mathbf{x_q}$. By subtracting the detected negative attribute $\mathbf{w_n}$ from the query embedding, we remove the negative attribute to avoid two visually contradictory attributes (*e.g.*, *sleeveless* and *long-sleeve*) hurting the retrieval performance. Eqn. 7 indicates that our method actually uses multimodal linguistic regularities [10] with automatic negative attribute detection.

Because the subspace networks are trained with a *none-of-above* class, it might predict that $\mathbf{x_q}$ does not have any attributes in concept $C$. In this case, our method degenerates to the baseline method.

## 5. Experimental Results and Discussions

### 5.1. Experiment Setup

**Clothing detection**. Some works have shown that using detected clothing segmentations instead of entire images can achieve better performance in various tasks [6, 8], so we also train a detector for each clothing category using MultiBox model [25] to detect and crop clothing items in our dataset. Because the product images on shopping websites have clean backgrounds, the detectors work very well.

**Visual-semantic embedding**. We use GoogleNet InceptionV3 model [26] for the image CNN. Its global average pooling (GAP) layer after the last convolutional layer enables us to directly use it without changing the structure of the network as in [34]. We use the 2048D features right after GAP as the image features. The dimension of the joint embedding space is set to 512, thus $\mathbf{W_I}$ is a $2048 \times 512$ matrix, and $\mathbf{W_T}$ is an $M \times 512$ matrix, where $M$ is the number of attributes. We set the margin $m = 0.2$ in Eqn. 1. The initial learning rate is 0.05 and is decayed by a factor of 2 after every 8 epochs. The batch size is set to 32. Finally, we fine-tune all layers of the network pretrained on ImageNet.

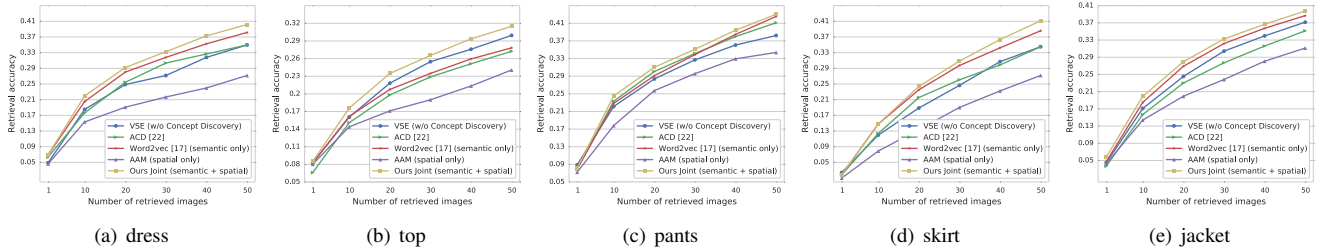

Figure 6. Top-k retrieval accuracy of different methods for attribute-feedback product retrieval for dresses, tops, pants, skirt, and jacket.

**Spatially-aware concept discovery.** The feature map size of the last convolutional layer in the InceptionV3 model is $8 \times 8 \times 2048$, hence the attribute activation map is of size $8 \times 8$. After vectorizing the activation map, an attribute will have a 64D feature vector as its spatial representation. We also set the dimension of word vectors to 64 to have the same dimentionality when training the Word2vec[18] model. The number of clusters is fixed to 50 for clustering.

**Subspace feature learning.** We set the hidden layer of each concept subspace to have 128 neurons. The learning rate is fixed to be 0.1 and we stop training after 10 epochs. Note that during training subspace networks, the visual-semantic embedding weights are fixed, only the parameters after the image embedding layer are updated.

### 5.2. Evaluation of Discovered Concepts

To evaluate the quality of our discovered concepts, a fashion professional manually assigned around 300 attributes into different categories (*e.g.*, *color*, *pattern*, *neckline*, *sleeve*, *etc.*). We use this information as ground truth concept assignments of the attributes and compare our approach with the following methods: Automatic Concept Discovery (ACD) [24], only using semantic representations of attributes for clustering (Word2vec[18]) and only using spatial information (Our AAM). In all methods, we set the number of clusters to 50. Homogeneity, completeness and V-measure [19] are used to evaluate the clustering quality.

Results are shown in Table 2. Only using semantic information gives reasonable results. However, just relying on spatial information performs worst, since for many attributes, their spatial information is not discriminative and thus fails to discover informative concepts. ACD performs similarly to Word2vec because it combines semantic and visual similarities of attributes but visually dissimilar at-tributes may also describe the same characteristic. By jointly clustering the semantic and spatial representations of attributes, our concept discovery approach outperforms other methods by 0.03 in V-measure.

|  | Homogeneity | Completeness | V-measure |
|---|---|---|---|
| ACD [24] | 0.770 | 0.527 | 0.626 |
| Word2vec | 0.765 | 0.534 | 0.629 |
| Ours AAM | 0.680 | 0.447 | 0.540 |
| Ours Joint | **0.794** | **0.561** | **0.658** |

Table 2. Comparison among concept discovery methods. Homogeneity, completeness and V-measure [19] are between 0 and 1, higher is better.

### 5.3. Attribute-feedback Product Retrieval

To evaluate how the discovered concepts help attribute-feedback product retrieval, we collected 3,167 product pairs from the test set. The two products in each pair have one attribute that differs in their product descriptions, *e.g.*, "blue geometric long-sleeve shirt" vs. "blue paisley long-sleeve shirt", "blue off-shoulder floral dress" vs. "blue one-shoulder floral dress", *etc*. In each pair, we use the image of one product and the differing attribute in their descriptions as the query to retrieve the images of the other product. Top-k retrieval accuracy is used for evaluation.

As shown in Figure 6, we compare our full method for all five categories with other methods. We also include the baseline method (VSE w/o concept discovery as in Eqn. 6), where no negative attribute is used.

We can see that using only attribute activation maps (AAM) significantly reduces performance of retrieval due to lack of semantic information. Only using semantic information (Word2vec) helps for most categories, but is worse than the baseline when retrieving tops. By adding visual information, ACD performs slightly worse than Word2vec because the visual similarity of attributes is not suitable for discovering concepts. After combining both semantic and spatial information, our concept discovery approach achieves the highest retrieval accuracy for all five categories, especially for the categories top, dress and jacket whose attributes have strong spatial information (*e.g.*, *collar shape*, *sleeve length*, *sleeve shape*). However, for clothing items like pants, whose attributes do not present informative spatial cues, our method only yields a marginal improvement over Word2vec.

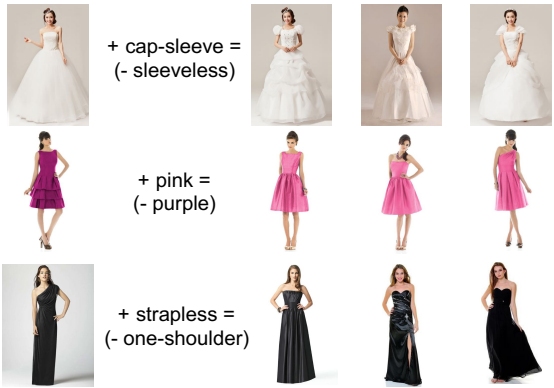

Figure 7. Examples of our attribute-feedback product retrieval results. *Sleeve type* changes from *sleeveless* to *cap-sleeve* in the first example, and *shoulder* changes from *one-shoulder* to *strapless* in the third example, according to customer feedback attributes. The attributes in parentheses are the negative attributes automatically detected by our method.

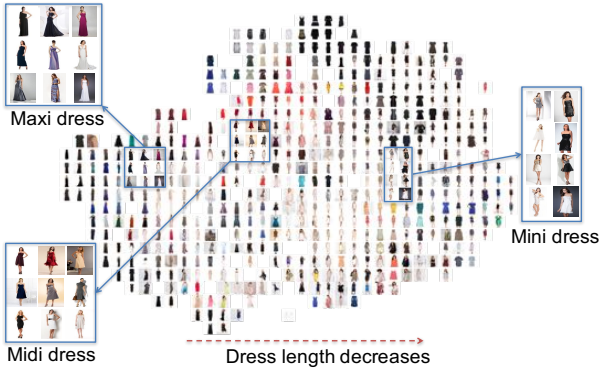

Figure 8. Subspace embedding corresponding to concept {maxi, midi, mini} for dresses. Images are mapped to a grid for better visualization.

Figure 7 illustrates some examples which show that our retrieval model can accurately detect the negative attribute and give satisfying results with the desired attributes added to the original results.

### 5.4. Structured Browsing of Products

Figure 8,9 use t-SNE [17] to visualize two subspace embeddings based on two discovered concepts. In Figure 8, the subspace network is trained to distinguish {maxi, midi, mini} for dresses, and it learns a continuous representation of the length of dresses - dress length decreases from left to right on the 2D visualization plane. Figure 9 illustrates the embedding corresponding to the attributes describing colors for tops. Tops with different colors are well separated in the embedding subspace. Although Veit *et al*. [29] also learns concept subspaces based on an attention mechanism, they heavily rely on richly annotated data, while our method is fully automatic and annotation free.

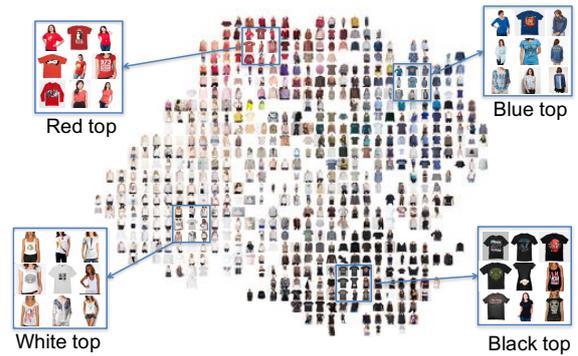

Figure 9. Subspace embedding corresponding to concept {black, blue, white, red, gray, green, purple, beige, ...} for tops.

By training a subspace embedding for each discovered concept, we can project images into the appropriate subspace and explore the images according to this specific concept, while a general embedding (like the visual-semantic embedding) cannot automatically adjust its representations based on user-specified characteristics.

Thus, the subspace features enable structured browsing during online shopping. For example, when a customer finds a mini dress and wants to see other dresses that share similar length with this dress, she may choose the subspace of {maxi, midi, mini}, so she can find the other mini dresses near her initial choice and as she explores the left side of the subspace, she can find dresses with longer length.

We should note that it is also possible to concatenate subspace embeddings of two concepts, hence clothing items sharing the same characteristics according to two concepts will be close in the concatenated subspace.

## 6. Conclusion

We automatically discover spatially-aware concepts with clothing images and their product descriptions. By projecting images and their attributes into a joint visual-semantic embedding space, we are able to learn attribute spatial representations. We then combine spatial representations and semantic representations of attributes, and cluster attributes into spatially-aware concepts, such that the attributes in one concept describe the same characteristic. Finally, a subspace embedding is trained for each concept to capture the concept-specific information. Experiments on clustering quality evaluation and attribute-feedback product retrieval for five clothing categories show the effectiveness of the discovered concepts and the learned subspace features.

## Acknowledgement

The authors acknowledge the Maryland Advanced Research Computing Center (MARCC) for providing computing resources.